\documentclass[conference]{IEEEtran}
\IEEEoverridecommandlockouts
\usepackage{cite}
\usepackage{amsmath,amssymb,amsfonts}
\usepackage{algorithmic}
\usepackage{graphicx}
\usepackage{textcomp}
\usepackage{bm}
\usepackage{amsmath}
\usepackage{breqn}
\usepackage{xcolor}

\usepackage{algorithm}

\usepackage{mathrsfs} 

\def\BibTeX{{\rm B\kern-.05em{\sc i\kern-.025em b}\kern-.08em
    T\kern-.1667em\lower.7ex\hbox{E}\kern-.125emX}}
\begin{document}

\title{Deep Approximately Orthogonal Nonnegative Matrix Factorization for Clustering}

\author{
\IEEEauthorblockN{Yuning Qiu}
\IEEEauthorblockA{\textit{School of Automation} \\
\textit{Guangdong University of Technology}\\
Guangzhou, China \\
yn.qiu@foxmail.com}
\and
\IEEEauthorblockN{Guoxu Zhou*}
\IEEEauthorblockA{\textit{School of Automation} \\
\textit{Guangdong University of Technology}\\
Guangzhou, China \\
guoxu.zhou@qq.com}
\and
\IEEEauthorblockN{Kan Xie}
\IEEEauthorblockA{\textit{School of Automation} \\
\textit{Guangdong University of Technology}\\
Guangzhou, China \\
kanxiegdut@gmail.com}
}

\maketitle

\begin{abstract}
Nonnegative Matrix Factorization (NMF) is a widely used technique for data representation. Inspired by the expressive power of deep learning, several NMF variants equipped with deep architectures have been proposed. However, these methods mostly use the only nonnegativity while ignoring task-specific features of data. In this paper, we propose a novel deep approximately orthogonal nonnegative matrix factorization method where both nonnegativity and orthogonality are imposed with the aim to perform a hierarchical clustering by using different level of abstractions of data. Experiment on two face image datasets showed that the proposed method achieved better clustering performance than other deep matrix factorization methods and state-of-the-art single layer NMF variants.
\end{abstract}

\begin{IEEEkeywords}
Deep Matrix Factorization,  Orthogonal NMF, Clustering Analysis
\end{IEEEkeywords}

\section{Introduction}
Nonnegative matrix factorization (NMF) is a useful tool for nonnegative data representation, with the capacity of preserving the non-negativity nature of data. NMF was first proposed by Paatero and Tapper \cite{paatero1994positive}, gained its popularity until Lee and Seung \cite{lee1999learning} discovered that it has the ability to learn part-based representation of objects. Over the past two decades, NMF has been found many successful applications in various areas such as text mining \cite{xu2003document}, \cite{pauca2004text}, \cite{shahnaz2006document}, speech separation \cite{tseng2015combining}, blind source separation \cite{zhou2015common} \cite{cichocki2009nonnegative}, and so on.

Orthogonal NMF, as one popular NMF variant, decomposes a nonnegative data matrix $\bm{X} \in \mathbb{R}^{M\times N}$ into nonnegative matrices $\bm{W}\in \mathbb{R}^{M\times R}$ and $\bm{H} \in \mathbb{R}^{N\times R}$ with constraints $\bm{H}^{T}\bm{H}=\bm{I}$, which can be achieved by solving the following optimization problem:
\begin{equation}\label{eq-onmf}
\min_{\bm{W}\geq 0,\bm{H}\geq 0} \parallel \bm{X} - \bm{W}\bm{H}^{T} \parallel_{F}^{2},~~s.t.\bm{H}^{T}\bm{H}=\bm{I}.
\end{equation}
Previous studies showed that standard NMF has clustering and pattern discovery effect empirically \cite{hoyer2004non} \cite{li2001learning}. Orthogonal NMF has been attached growing importance as it can be equivalent to the K-means method, where the column vectors of matrix $\bm{W}$  and $\bm{H}$ of orthogonal NMF just correspond to clustering centroids and category indicators, respectively \cite{ding2005equivalence}. Moreover, ONMF could minimize the redundancy between bases and correspond to a unique sparse area in the solution region, which learns the most distinct parts \cite{wang2013nonnegative}. 
For this reason, several efficient and robust orthogonal NMF algorithms have been developed \cite{li2015two} \cite{gillis2012accelerated}. 

Recently,  Deep Neural Network (DNN) is being studied extensively. Equipped with a multi-layer structure, DNN is able to learn lower dimensional higher-level representations of data, therefore provide unprecedented performance in many machine learning tasks. Inspired by the philosophy of DNN, many novel matrix factorization models equiped with deep architectures have been proposed. A nonnegative deep network based on NMF was proposed in \cite{le2015deep} and was applied to speech separation, and \cite{trigeorgis2017deep} set up Deep Semi-NMF and Weakly Supervised Deep Semi-NMF model to learn the latent representations by further factorizing feature matrices. \cite{zhao2017multi} presented a deep matrix factorization model and achieved satisfactory results for multi-view clustering. However, to the best of our knowledge, all of these kind of methods are derived from the idea of decomposing matrix $\bm{H}$ hierarchically in Multi-NMF \cite{cichocki2006multilayer}. One can suppose that by further decomposing the mapping matrix $\bm{W}$ hierarchically and fine-tuning the model, we are also able to obtain the higher-level representations for cluster. This paper extends upon our recent paper \cite{benshenglv}, with new algorithms and detailed derivations.

$\mathbf{Main~constributions}$: a novel deep nonnegative matrix factorization method is built by further factorizing mapping matrices. Based on this novel deep matrix factorizing technique, we proposed deep approximately orthogonal nonnegative matrix factorization (DAONMF) by incorporating the orthogonality penalty on each layer. After fine-tuning, DAONMF is able to learn the higher-level representations of data and therefore achieved remarkable clustering performance in two face image datasets.
\section{Background}

\subsection{HALS algorithm for AONMF}
Different from the conventional ONMF, approximately orthogonal NMF (AONMF) treats the orthogonality constrained of matrix $\bm{H}$ as a penalty term, which is added to the Lagrangian multiplier function to control the orthogonal strength, and has achieved satisfying clustering results \cite{pompili2014two} \cite{li2015two}. The cost function of AONMF can be formulated as:
\begin{equation}\label{eq-AONMF}
f(\bm{W},\bm{H}) = \frac{1}{2}\parallel \bm{X} - \bm{W}\bm{H}^{T}\parallel_{F}^{2} + \frac{\lambda}{2}\sum_{r=1}^{R} \sum_{j\neq r} {h_{r}^{T}h_{j}},
\end{equation}
where $\lambda$ controls the orthogonal degree of $\bm{H}^{T}$, and $\lambda \to +\infty$, $\bm{H}^{T}\bm{H} \to \bm{I}$. Since NMF is a non-convex optimization problem, it is extremely significant to locate the local convergence points. \cite{li2015two} proposed the Hierarchical Alternating Lease square (HALS) method for optimizing AONMF based on block coordinate descent (BCD) method \cite{bertsekas1999nonlinear}, which can achieve local convergence due to the optimal solutions of subproblems are always available  \cite{li2015two}. The update rule of HALS algorithm can be given as:
\begin{equation}\label{eq-hals-h}
\bm{h}_{r} \leftarrow \mathcal{P}_{+}(\bm{h}_{r} + \frac{\bm{X}^{T}\bm{w}_{r} - \bm{H}(\bm{W}^{T}\bm{w}_{r})}{\bm{w}_{r}^{T}\bm{w}_{r}} - \frac{\lambda\breve{\bm{H}}_{r}\bm{1}_{R-1}}{\bm{w}_{r}^{T}\bm{w}_{r}})
\end{equation}
and
\begin{equation}\label{eq-hals-w}
	\bm{w}_{r} \leftarrow \mathcal{P}_{+}(\bm{w}_{r} + \frac{\bm{X}\bm{h}_{r}-\bm{W}(\bm{H}^{T}\bm{h}_{r})}{\bm{h}^{T}\bm{h}})
\end{equation}
Where $\mathcal{P}_{+}(\bm{H}) $ is the nonnegative projection of $\bm{\bm{H}}$, $\bm{h}_{r}$ and $\bm{w}_{r}$ are the $r$th column vectors with respect to matrix $\bm{H}$ and $\bm{W}$, $\bm{1}_{R-1}$ is an all-one R-by-1 column vector, $\breve{\bm{H}}$ is the the sub-matrix with $\bm{H}$ removed the $r$th column vector. Since the column vector $\bm{h}_{r}$ can not be zero, each sub-problem with respect to column $\bm{h}_{r}$ will converge to their global minimum, which is equivalent to the matrix $\bm{H}$ with orthogonality constraint, we can always attain its unique optimal solution \cite{li2015two}.

\subsection{Deep Semi-NMF}

The single layer NMF structure indicates intuitively that $\bm{X}$ is approximately equivalent to the linear additive between $\bm{W}$ and $\bm{H}$, thus resulting in the failure of learning more abstract features of $\bm{X}$. The Deep Semi-NMF \cite{trigeorgis2017deep}, a hierarchical matrix factorization model, aims to learn a hierarchical features for original data, which can be defined as follow:
\begin{equation}\label{eq-dsemi}
\bm{X} \approx \bm{W}_{1}^{\pm}\bm{W}_{2}^{\pm}...\bm{W}_{L}^{\pm}\bm{H}_{L}^{+},
\end{equation}
To avoid getting stuck in poor solution, the layer wise pre-taring technique, which is widely used in deep neural network \cite{bengio2007greedy}, has been adopted for initializing $\bm{W}_{l}$ and $\bm{H}_{l}$. For each layer, $\bm{H}_{l-1}^{+}$ is decomposed into $\bm{W}_{l}{}^{\pm}$ and $\bm{H}_{l}^{+}$ by implementing Semi-NMF method. The Deep Semi-NMF model is initialized when $\bm{X}$ is decomposed into $m+1$ factors. Due to the equivalence between semi-NMF and K-means clustering \cite{ding2005equivalence} \cite{ding2010convex} and multi-layer structure, Deep Semi-NMF learned the hierarchical projections from data space to subspaces spanned by the hidden attributes, therefore achieved remarkable clustering results \cite{trigeorgis2017deep}.

\section{Deep Approximately Orthogonal NMF}
\subsection{Model Setup}
The DAONMF model factorizes the nonnegative data matrix $\bm{X}$ into $L+1$ nonnegative factor matrices:
\begin{equation}\label{eq-DAONMF}
	\bm{X} \approx \bm{W}_{1}\bm{H}_{1}^{T} \cdot \cdot \cdot \bm{H}_{L}^{T},
\end{equation}
An intuitive representation of the hierarchical factorization can be shown as follow:
\begin{equation}\label{eq-ManW}
\begin{aligned}
\bm{X} \approx \bm{W}_{L} \bm{H}_{L}^{T},\\
\bm{W}_{L}\approx \bm{W}_{L-1} \bm{H}_{L-1}^{T},\\
\cdot\cdot\cdot, \\
\bm{W}_{3}\approx \bm{W}_{2} \bm{H}_{2}^{T},\\
\bm{W}_{2}\approx \bm{W}_{1} \bm{H}_{1}^{T},
\end{aligned}
\end{equation}
Different from other multi-layer matrix factorization methods  \cite{le2015deep}, \cite{trigeorgis2017deep}, \cite{zhao2017multi}, \cite{cichocki2006multilayer}, the proposed method is built by further factorizing $\bm{W}_{l}$ instead of $\bm{H}_{l}$. We assume that by incorporating fine-tuning technique, features learned by DAONMF are also beneficial to the clustering analysis.  As illustrated in (\ref{eq-ManW}), the original data matrix $\bm{X}$ is decomposed into $\bm{W}_{L}\bm{H}_{L}^{T}$, then $\bm{W}_{L}$ will be further factorized into second layer $\bm{W}_{L-1} \bm{H}_{L-1}^{T}$, and the rest layers can be done in the same manner. After the hierarchical decomposition of $\bm{W}_{l}$, a deep NMF model will be initialized. Previous studies have shown that, NMF with approximate orthogonality constrained is able to improve clustering accuracy \cite{li2015two}. For this reason, orthogonality constraint is adopted to $\bm{H}_{l}^{T}$ of each layer, therefore denoted as Deep Approximately Orthogonal NMF (DAONMF). The objective function can be formulated as follows:
\begin{align}\label{eq-obj}
\begin{split}
\min \quad &\frac{1}{2}\parallel\bm{X}-\bm{W}_{1}\bm{H}_{1}^{T}\cdot\cdot\cdot\bm{H}_{L}^{T}\parallel_{F}^{2}\\ 
&+ \frac{\lambda}{2}\sum_{l=1}^{L} \sum_{r=1}^{R}\sum_{j \neq r}{\bm{H}_{l}(:,r)^{T}\bm{H}_{l}(:,j)}, \\
\text{s.t.}  \quad & \bm{W}_{l}\geq 0, \bm{H}_{l} \geq 0,
\end{split}
\end{align}
where (\ref{eq-obj}) is an extension of (\ref{eq-AONMF}) with multiple approximately orthogonal matrices. It is clear that $\bm{H}_{L}^{T}$ is attained at the time of initialization and is exactly equal to that of AONMF in (\ref{eq-AONMF}), which means that in this case $\bm{H}_{L-1}^{T}$ is complementary useless to $\bm{H}_{L}^{T}$. However, as the model is fine-tuning, $\bm{H}_{L}^{T}$ will be able to learn the universe features from the later factorized layers, and therefore achieve better clustering.
\subsection{Optimization}
As shown in  (\ref{eq-DAONMF}), the initialized DAONMF model is obtained by further factorizing $\bm{H}_{l}$. Afterwards, the fine-tuning technique is implemented in order to learn the universal features factorized from the later layers. The cost function can be denoted as:
\begin{align}\label{eq-cost}
\begin{split}
Cost=&\frac{1}{2}\parallel \bm{X} - \bm{W}_{1}\bm{H}_{1}^{T} \cdot \cdot \cdot \bm{H}_{L-1}^{T}\bm{H}_{L}^{T}\parallel _{F}^{2} \\
&+ \frac{\lambda}{2}\text{tr}(\bm{H}_{L}\bm{Q}\bm{H}_{L}^{T}).
\end{split}
\end{align}

\paragraph{Update rule for matrix $\bm{H}_{L}$} $\bm{H}_{L}$ is a nonnegative approximately orthogonal matrix, therefore we can formulate the update rule for $\bm{H}_{L}$ utilizing HALS algorithm:
\begin{align}\label{eq-UpHL}
\begin{split}
(\bm{H}_{L})_{r} \leftarrow \mathcal{P}_{+}&((\bm{H}_{L})_{r} + \frac{\bm{X}^{T}(\bm{W}_L)_{r} - \bm{H}_{L}(\bm{W}_{L}^{T}(\bm{W}_L)_{r})}{(\bm{W}_L)_{r}^{T}(\bm{W}_L)_{r}} \\
& - \frac{\lambda(\breve{\bm{H}}_{L})_{r}\bm{1}_{R-1}}{(\bm{W}_L)_{r}^{T}(\bm{W}_L)_{r}}),
\end{split}
\end{align}
where $(\bm{H}_{L})_{r}$ and $(\bm{W}_{L})_{r}$ are the $r$th columns vector of $\bm{H}_{L}$ and $\bm{W}_{L}$ respectively, $(\breve{\bm{H}}_{L})_{r}$ is a sub-matrix of $\bm{H}_{L}$ with removing the $r$th column vector, $\bm{1}_{R-1}$ is an all-one (R-1)-by-1 column vector and R is equal to the number of column vector of $\bm{H}_{L}$. 

\paragraph{Update rule for matrix $\bm{W}_{1}$} By incorporating the multiplicative update rules proposed in \cite{lee2001algorithms}, we can formulate it as:
\begin{equation}\label{eq-UpW1}
(\bm{W}_{1})_{jk} \leftarrow (\bm{W}_{1})_{jk} \frac{(\bm{X}\bm{\varPhi}^{T})_{jk}}{(\bm{W}_{1} \bm{\varPhi}\bm{\varPhi}^{T})_{jk}},
\end{equation}
where $\bm{\varPhi} = \bm{H}_{1}^{T}\bm{H}_{2}^{T}\cdot \cdot \cdot \bm{H}_{L}^{T}$.

\paragraph{Update rule for matrices $\bm{H}_{l}$ (l=1,2,...,L-1)} Since $\bm{H}_{l}$ can not be updated like  (\ref{eq-UpHL}) and (\ref{eq-UpW1}). Therefore, for the ease of mathematical, we represent the cost function with another form as:
\begin{equation}
Cost(\bm{H}_{l}^{T})= \frac{1}{2}\parallel \bm{X} - \bm{W}_{l} \bm{H}_{l}^{T}\bm{ \varPsi} \parallel_{F}^{2} +  \frac{\lambda}{2}\bm{1}_{R}^{T}(\bm{H}_{l}^{T}\bm{H}_{l} - \bm{I})\bm{1}_{R}
\end{equation}
where $\bm{\varPsi}_{l} = \prod_{i=l+1}^{L} \bm{H}_{i}$. The gradient of $\bm{H}_{l}$ is solved as:
\begin{equation}\label{eq-gradient}
\nabla_{\bm{H}_{l}} = \bm{\varPsi}^{T}\bm{\varPsi}\bm{H}_{l}^{T} \bm{W}_{l}\bm{W}_{l}^{T}  - \bm{\varPsi}\bm{X} ^{T}\bm{W}_{l} + \lambda \bm{H}_{l} \bm{1}_{R\times R}^{T}
\end{equation}
thus the additive update for $\bm{W}_{l}$ can be given as:
\begin{equation}\label{eq-gradientH}
(\bm{H}_{l})_{jk} \leftarrow (\bm{H}_{l})_{jk} - \eta_{jk}(\nabla_{\bm{H}_{l}})_{jk}
\end{equation}
Inspired by \cite{lee2001algorithms}, multiplicative update rule is solved if $\eta_{jk} = \frac{(\bm{H}_{l})_{jk}}{{(\bm{W}_{l}^{T}\bm{W}_{l}\bm{H}_{l}^{T}\bm{\varPsi}\bm{\varPsi}^{T}}+\lambda \bm{1}_{R\times R}\bm{H}_{l}^{T})_{jk}}$, therefore the convergence is guaranteed.
The proposed multiplicative update rule is given as:
\begin{equation}\label{eq-Hi}
(\bm{H}_{l})_{jk} \leftarrow (\bm{H}_{l})_{jk} \frac{(\bm{\varPsi}\bm{X}^{T}\bm{W}_{l})_{jk}}{(\bm{\varPsi}^{T}\bm{\varPsi}\bm{H}_{l}\bm{W}_{l}\bm{W}_{l}^{T} + \lambda  \bm{H}_{l} \bm{1}_{R\times R})_{jk}}
\end{equation}

We have presented the update rules of each variables in DAONMF model above, and the entire optimization algorithm process is outlined in $\mathbf{Algorithm~1}$.

\begin{algorithm}
 \caption{Algorithm for training a Deep Approximately Orthogonal NMF (DAONMF)}
 \begin{algorithmic}[1]
 \renewcommand{\algorithmicrequire}{\textbf{Input:}}
 \renewcommand{\algorithmicensure}{\textbf{Output:}}
 \REQUIRE $\bm{X}\in \mathbb{R}^{M\times N}$, layer size array $d$, fine-tuning parameters $\lambda$ 
 \ENSURE  Matrices $\bm{W}$ and $\bm{H}^{T}$
 \\ \textit{Initialisation process} :
  \FOR {$l = L$ to $1$}
    \STATE $\bm{W}_{l},\bm{H}_{l} \leftarrow$ AONMF($\bm{H}_{l+1}$, $d_{l}$)
  \ENDFOR
 \STATE
\textit{Fine-tuning process} :
 \REPEAT
   \FOR {$l=1$ to $L$}
  \STATE $\bm{\varPhi} \leftarrow \prod_{i=1}^{L}\bm{H}_{i}^{T}$
  \STATE $\bm{\varPsi} \leftarrow \prod_{i=l+1}^{L} \bm{H}_{i}^{T}$
  \STATE $\bm{W}_{1} \leftarrow$ Update via (\ref{eq-UpW1})
\STATE $\bm{W}_{l} \leftarrow \bm{W}_{1}\prod_{i=1}^{l-1}\bm{H}_{i}$

   \IF {$l < L$}
   \STATE Update $\bm{H}_{l}$ using (\ref{eq-Hi})
     
   \ELSE
     \FOR {all columns}
    	\STATE Update $(\bm{H}_{L})_{r}$ using (\ref{eq-UpHL})
    \ENDFOR
  
   \ENDIF
    	
   \ENDFOR
   \STATE Adjust the Columns of $\bm{H}_{L}$ such that $\parallel (\bm{H}_{L})_{r} \parallel_{2}= 1$
 \UNTIL  Stopping criterion is reached

 \end{algorithmic} 
 \end{algorithm}

\section{Experiments}
In order to demonstrate the effectiveness of DAONMF, we have taken comparative experiments on two different face images datasets. The methods compared include state-of-art single layer NMF variants NeNMF \cite{guan2012nenmf} and AONMF\cite{li2015two}, and Deep Semi-NMF \cite{trigeorgis2017deep}. To ensure the evaluations to be comprehensive, we have adopted two evaluation criteria including clustering accuracy (ACC) and normalized mutual information (NMI) \cite{cai2011graph}.  
\subsection{Datasets }
$\mathbf{CMU~PIE:}$ The CMU PIE dataset \cite{sim2002cmu} contains 2856 frontal-face images of 68 people, and each object has 42 facial images different illumination. In this paper, each image was cropped into $32\times 32$ pixels.

$\mathbf{Extended~Yale~B:}$ The Extended Yale B dataset \cite{KCLee05} comprises of 16128 images of 28 people under 9 poses and 64 lighting conditions. We used the front pose face images of the first 10 objects and cropped the $32\times 32$ pixels.

\subsection{Implementation Details}
Similar to the experimental procedure in \cite{trigeorgis2017deep}, we also set the number of layers $L = 2$. On the CMU PIE dataset, the following parameters are selected. The number of the first layer $k_{1} = 120$ and the second layer $k_{2} = 130,135,140,145,150,160,165,170$ are fixed for DAONMF.  As for Deep Semi-NMF, we chose $k_{1}^{'}=600$ \cite{trigeorgis2017deep} for the first layer and $k_{2}$ for the second layer. And the number of components of single layer methods ONMF and NeNMF is also set to $k_{2}$. Since orthogonality constraints is significant to clustering, we chose $\lambda_{1}=1e-6$ for the first layer, and $\lambda_{2}=1e-5$ for the second layer and AONMF. After the decompositions, K-means clustering is implemented to the factorized features. On the Extended Yale B dataset, we perform exactly the same operation, with $k_{2}$ changing to $110,115,120,125,130,135,140,145,150 $ and $k_{1}$ replacing to $100$.

\begin{figure}[h]
\centering
 \includegraphics[scale=0.54]{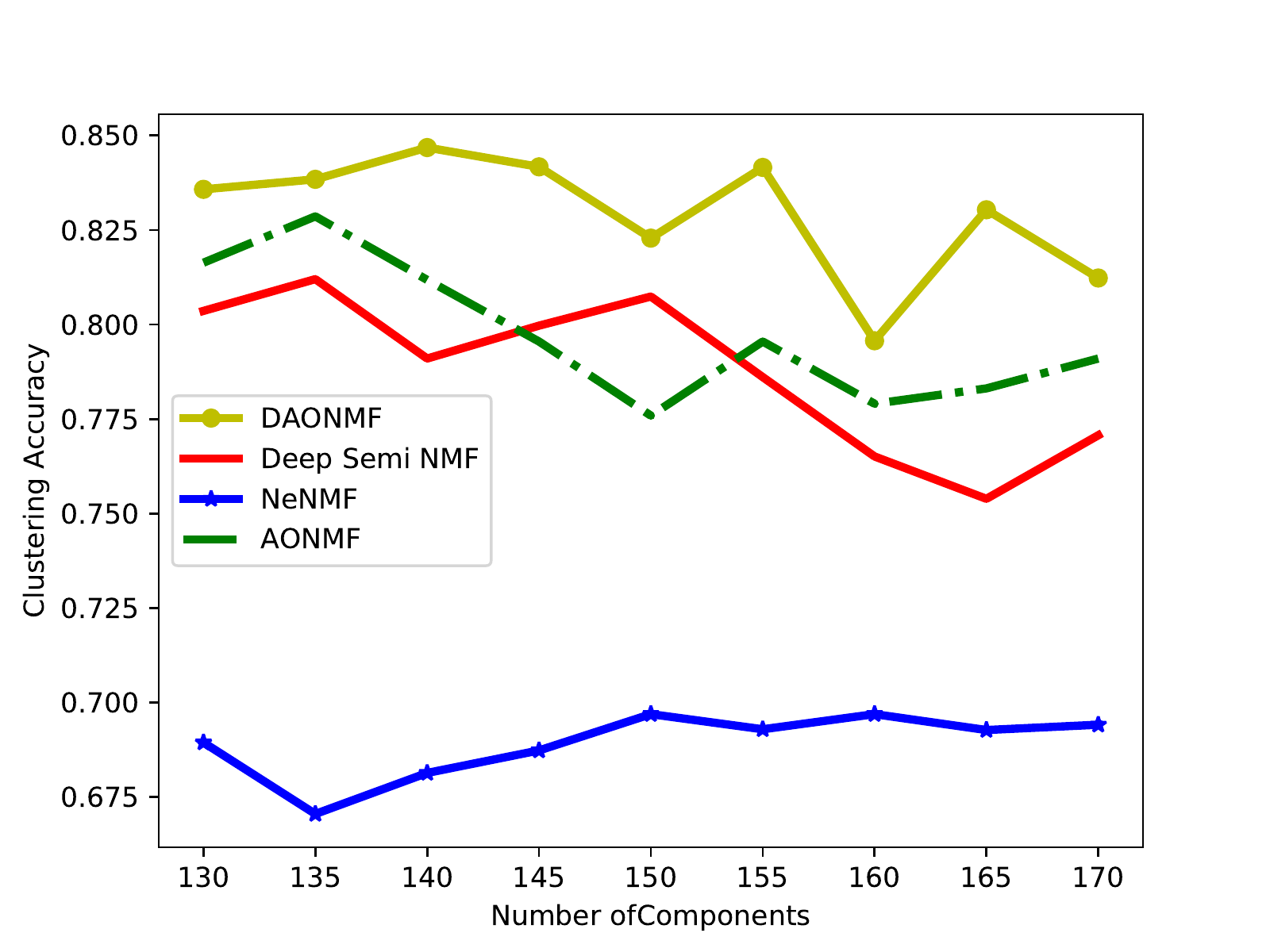}
 \caption{Accuracy for clustering on dataset CMU PIE}
 \label{fig-cmuacc}
\end{figure}

\begin{figure}
\centering
 \includegraphics[scale=0.54]{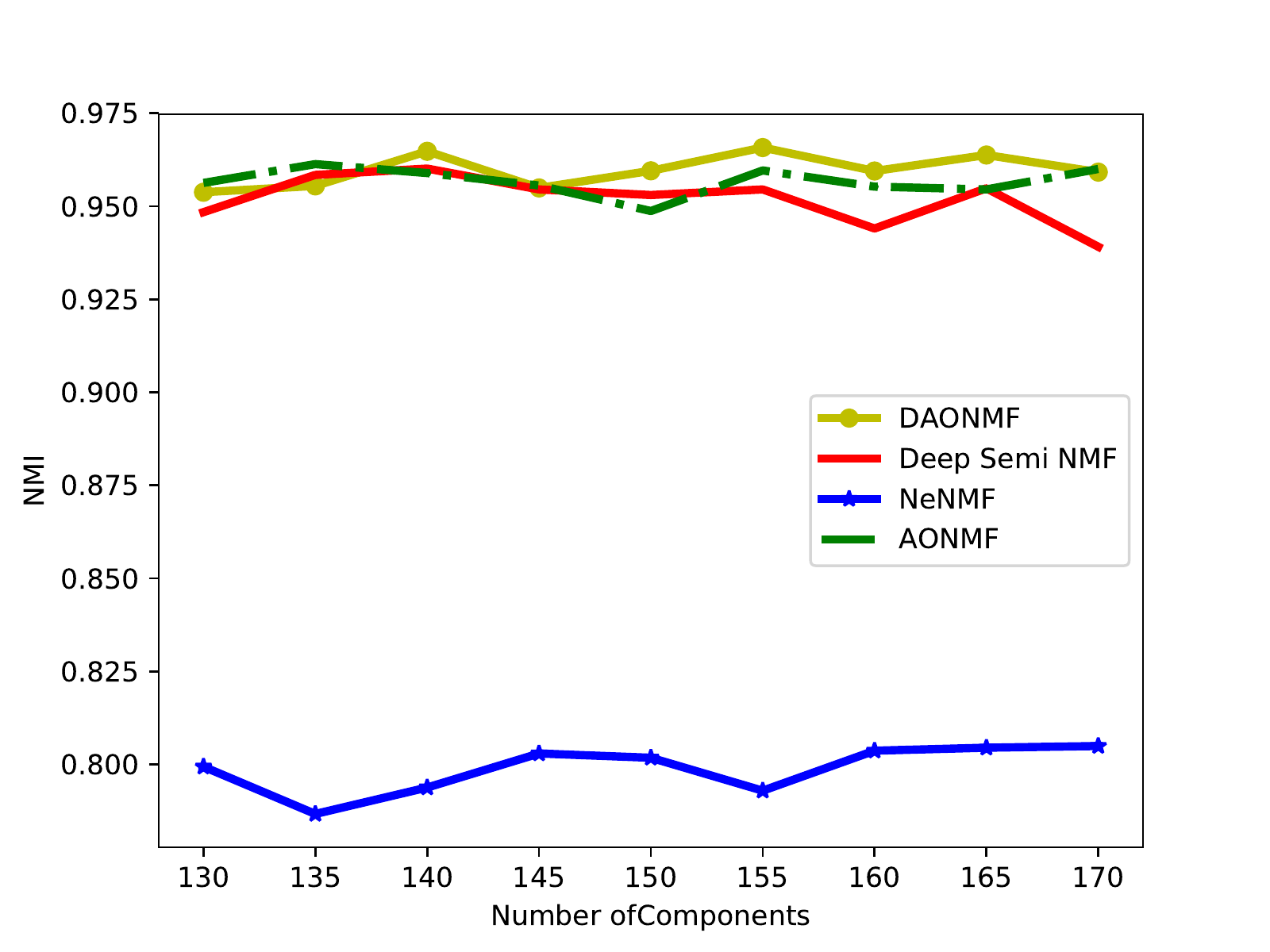}
 \caption{NMI for clustering on dataset CMU PIE}
 \label{fig-cmunmi}
\end{figure}

\begin{figure}
\centering
 \includegraphics[scale=0.54]{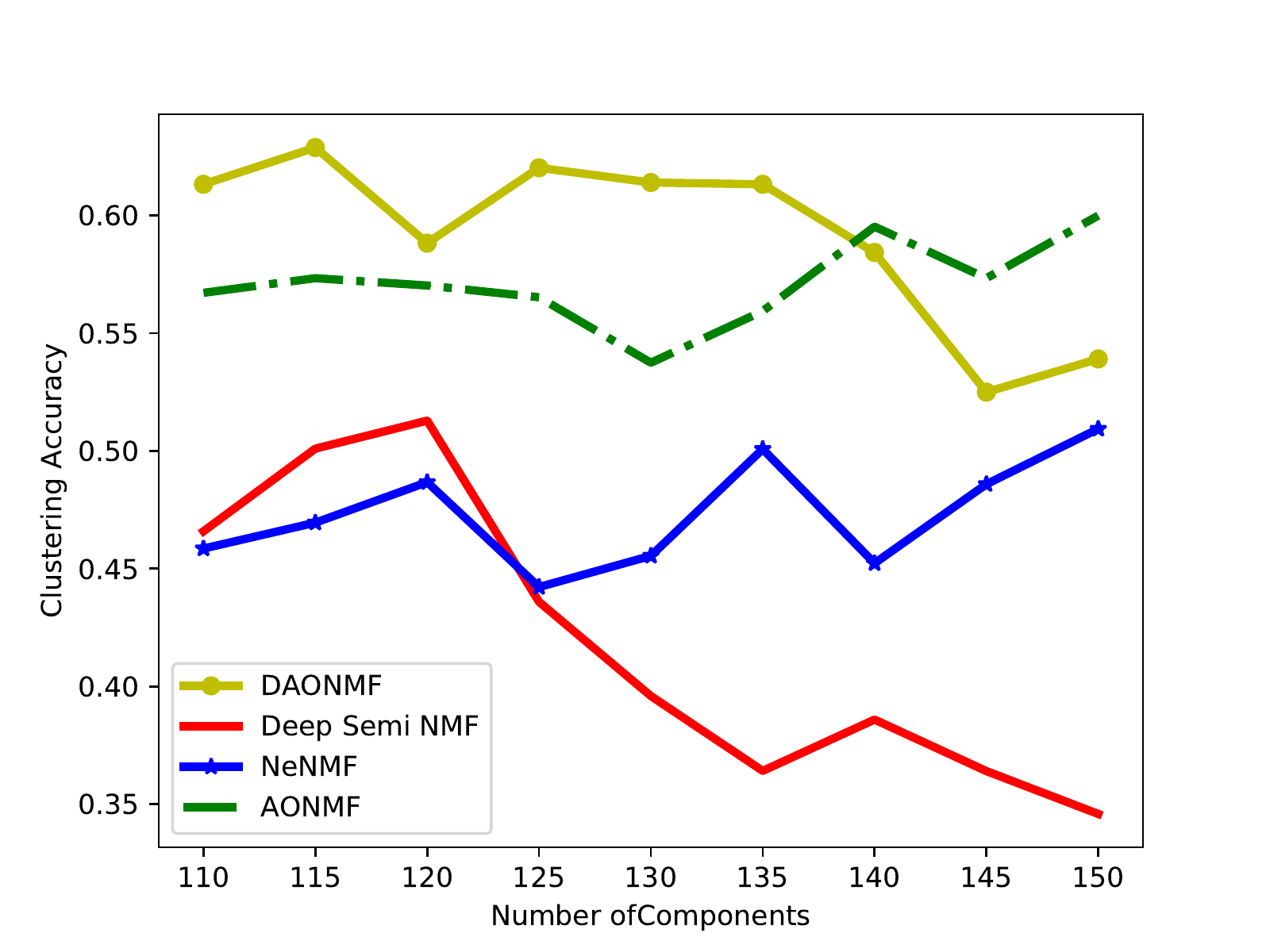}
 \caption{Accuracy for clustering on dataset Extended Yale B}
 \label{fig-yaleacc}
\end{figure}

\begin{figure}
\centering
 \includegraphics[scale=0.54]{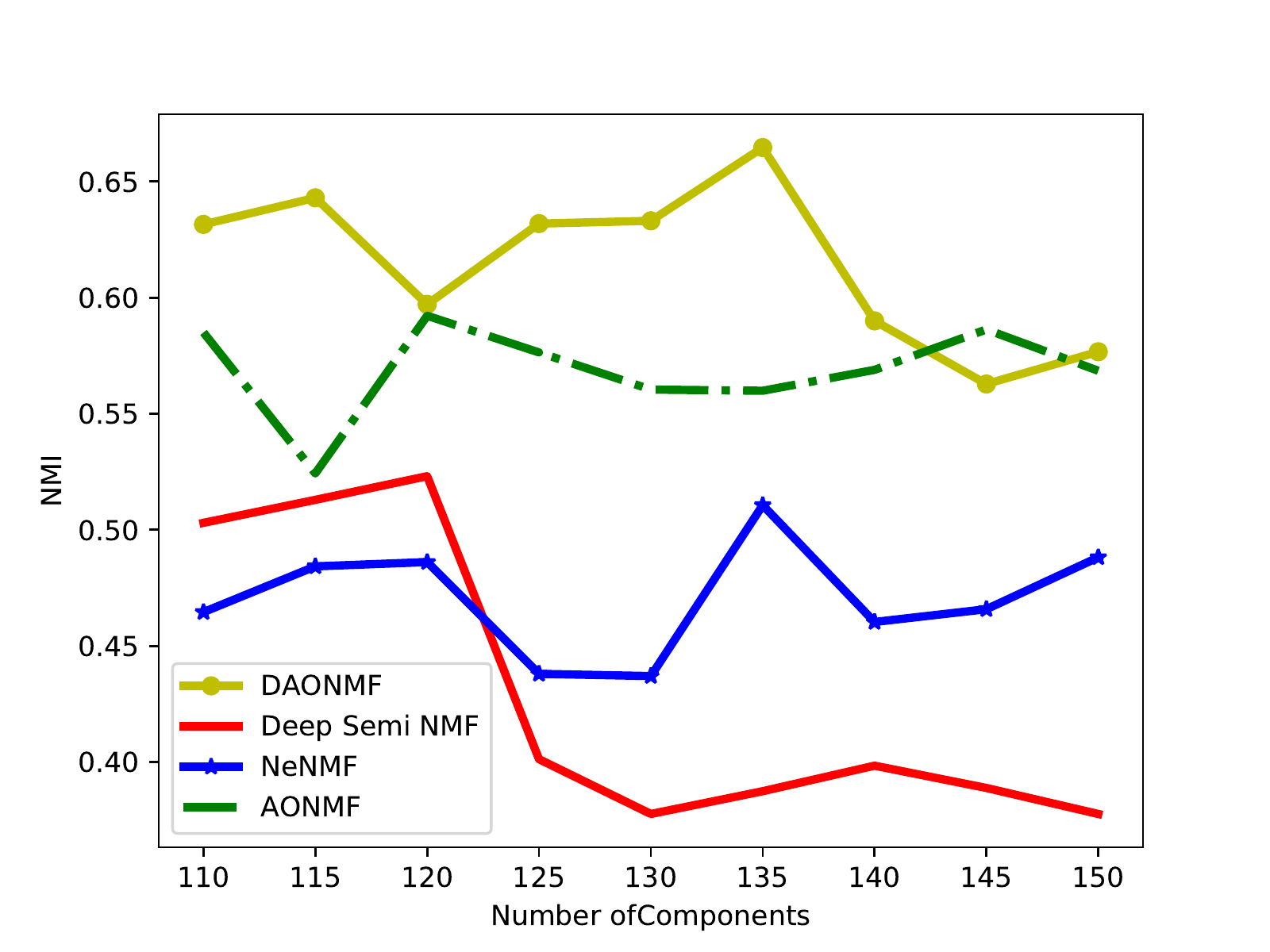}
 \caption{NMI for clustering on dataset Extended Yale B}
 \label{fig-yalenmi}
\end{figure}

\subsection{Results and Discussion}

%
Figure \ref{fig-cmuacc} to \ref{fig-yalenmi} demonstrate clustering accuracy and NMI for CMU PIE and Extended Yale B dataset. DAONMF yields better clustering accuracy and NMI than single layer AONMF, which implies that DAONMF learns higher-level features from later factorized layers by fine-tuning. 
The single layer AONMF achieves even better clustering performance than Deep Semi-NMF. Therefore in some cases the single layer AONMF with sparse and low-redundancy representations are more efficient than Deep Semi-NMF in clustering analysis.
Figure \ref{fig-yaleacc} and \ref{fig-yalenmi} present that both clustering accuracy and NMI of multi-layer NMFs decrease with the increases of the number of clustering components, but the single layer NMFs are the opposite. This is mainly because multi-layer NMFs are able to factorize higher-level features with lower dimensional representations, thus the performance will be reduced with redundant features. 

\section{Conclusion}
In this paper, we proposed a novel deep approximately orthogonal nonnegative matrix factorization (DAONMF) method for clustering analysis. Different from other deep (semi-) nonnegative matrix factorization methods, the proposed method incorporates both nonnegativity and orthogonality into a deep architecture. Experiments confirmed that the proposed method is able to extract higher-level lower dimensional sparse features that is particularly beneficial for clustering analysis.


\bibliographystyle{IEEEtran}
\bibliography{ref.bib}

\begin{thebibliography}{10}
\providecommand{\url}[1]{#1}
\csname url@samestyle\endcsname
\providecommand{\newblock}{\relax}
\providecommand{\bibinfo}[2]{#2}
\providecommand{\BIBentrySTDinterwordspacing}{\spaceskip=0pt\relax}
\providecommand{\BIBentryALTinterwordstretchfactor}{4}
\providecommand{\BIBentryALTinterwordspacing}{\spaceskip=\fontdimen2\font plus
\BIBentryALTinterwordstretchfactor\fontdimen3\font minus
  \fontdimen4\font\relax}
\providecommand{\BIBforeignlanguage}[2]{{%
\expandafter\ifx\csname l@#1\endcsname\relax
\typeout{** WARNING: IEEEtran.bst: No hyphenation pattern has been}%
\typeout{** loaded for the language `#1'. Using the pattern for}%
\typeout{** the default language instead.}%
\else
\language=\csname l@#1\endcsname
\fi
#2}}
\providecommand{\BIBdecl}{\relax}
\BIBdecl

\bibitem{paatero1994positive}
P.~Paatero and U.~Tapper, ``Positive matrix factorization: A non-negative
  factor model with optimal utilization of error estimates of data values,''
  \emph{Environmetrics}, vol.~5, no.~2, pp. 111--126, 1994.

\bibitem{lee1999learning}
D.~D. Lee and H.~S. Seung, ``Learning the parts of objects by non-negative
  matrix factorization,'' \emph{Nature}, vol. 401, no. 6755, p. 788, 1999.

\bibitem{xu2003document}
W.~Xu, X.~Liu, and Y.~Gong, ``Document clustering based on non-negative matrix
  factorization,'' in \emph{Proceedings of the 26th annual international ACM
  SIGIR conference on Research and development in informaion retrieval}.\hskip
  1em plus 0.5em minus 0.4em\relax ACM, 2003, pp. 267--273.

\bibitem{pauca2004text}
V.~P. Pauca, F.~Shahnaz, M.~W. Berry, and R.~J. Plemmons, ``Text mining using
  non-negative matrix factorizations,'' in \emph{Proceedings of the 2004 SIAM
  International Conference on Data Mining}.\hskip 1em plus 0.5em minus
  0.4em\relax SIAM, 2004, pp. 452--456.

\bibitem{shahnaz2006document}
F.~Shahnaz, M.~W. Berry, V.~P. Pauca, and R.~J. Plemmons, ``Document clustering
  using nonnegative matrix factorization,'' \emph{Information Processing \&
  Management}, vol.~42, no.~2, pp. 373--386, 2006.

\bibitem{tseng2015combining}
H.-W. Tseng, M.~Hong, and Z.-Q. Luo, ``Combining sparse nmf with deep neural
  network: A new classification-based approach for speech enhancement,'' in
  \emph{Acoustics, Speech and Signal Processing (ICASSP), 2015 IEEE
  International Conference on}.\hskip 1em plus 0.5em minus 0.4em\relax IEEE,
  2015, pp. 2145--2149.

\bibitem{zhou2015common}
G.~Zhou, A.~Cichocki, and D.~P. Mandic, ``Common components analysis via linked
  blind source separation,'' in \emph{Acoustics, Speech and Signal Processing
  (ICASSP), 2015 IEEE International Conference on}.\hskip 1em plus 0.5em minus
  0.4em\relax IEEE, 2015, pp. 2150--2154.

\bibitem{cichocki2009nonnegative}
A.~Cichocki, R.~Zdunek, A.~H. Phan, and S.-i. Amari, \emph{Nonnegative matrix
  and tensor factorizations: applications to exploratory multi-way data
  analysis and blind source separation}.\hskip 1em plus 0.5em minus 0.4em\relax
  John Wiley \& Sons, 2009.

\bibitem{hoyer2004non}
P.~O. Hoyer, ``Non-negative matrix factorization with sparseness constraints,''
  \emph{Journal of machine learning research}, vol.~5, no. Nov, pp. 1457--1469,
  2004.

\bibitem{li2001learning}
S.~Z. Li, X.~W. Hou, H.~J. Zhang, and Q.~S. Cheng, ``Learning spatially
  localized, parts-based representation,'' in \emph{Computer Vision and Pattern
  Recognition, 2001. CVPR 2001. Proceedings of the 2001 IEEE Computer Society
  Conference on}, vol.~1.\hskip 1em plus 0.5em minus 0.4em\relax IEEE, 2001,
  pp. I--I.

\bibitem{ding2005equivalence}
C.~Ding, X.~He, and H.~D. Simon, ``On the equivalence of nonnegative matrix
  factorization and spectral clustering,'' in \emph{Proceedings of the 2005
  SIAM International Conference on Data Mining}.\hskip 1em plus 0.5em minus
  0.4em\relax SIAM, 2005, pp. 606--610.

\bibitem{wang2013nonnegative}
Y.-X. Wang and Y.-J. Zhang, ``Nonnegative matrix factorization: A comprehensive
  review,'' \emph{IEEE Transactions on Knowledge and Data Engineering},
  vol.~25, no.~6, pp. 1336--1353, 2013.

\bibitem{li2015two}
B.~Li, G.~Zhou, and A.~Cichocki, ``Two efficient algorithms for approximately
  orthogonal nonnegative matrix factorization,'' \emph{IEEE Signal Processing
  Letters}, vol.~22, no.~7, pp. 843--846, 2015.

\bibitem{gillis2012accelerated}
N.~Gillis and F.~Glineur, ``Accelerated multiplicative updates and hierarchical
  als algorithms for nonnegative matrix factorization,'' \emph{Neural
  computation}, vol.~24, no.~4, pp. 1085--1105, 2012.

\bibitem{le2015deep}
J.~Le~Roux, J.~R. Hershey, and F.~Weninger, ``Deep nmf for speech separation,''
  in \emph{Acoustics, Speech and Signal Processing (ICASSP), 2015 IEEE
  International Conference on}.\hskip 1em plus 0.5em minus 0.4em\relax IEEE,
  2015, pp. 66--70.

\bibitem{trigeorgis2017deep}
G.~Trigeorgis, K.~Bousmalis, S.~Zafeiriou, and B.~W. Schuller, ``A deep matrix
  factorization method for learning attribute representations,'' \emph{IEEE
  transactions on pattern analysis and machine intelligence}, vol.~39, no.~3,
  pp. 417--429, 2017.

\bibitem{zhao2017multi}
H.~Zhao, Z.~Ding, and Y.~Fu, ``Multi-view clustering via deep matrix
  factorization.'' in \emph{AAAI}, 2017, pp. 2921--2927.

\bibitem{cichocki2006multilayer}
A.~Cichocki and R.~Zdunek, ``Multilayer nonnegative matrix factorisation,''
  \emph{Electronics Letters}, vol.~42, no.~16, p.~1, 2006.

\bibitem{benshenglv}
\BIBentryALTinterwordspacing
B.~Lyu, K.~Xie, and W.~Sun, ``A deep orthogonal non-negative matrix
  factorization method for learning attribute representations,'' in
  \emph{International Conference On Neural Information Processing}.\hskip 1em
  plus 0.5em minus 0.4em\relax Springer, 2017. [Online]. Available:
  \url{https://doi.org/10.1007/978-3-319-70136-3\textunderscore47}
\BIBentrySTDinterwordspacing

\bibitem{pompili2014two}
F.~Pompili, N.~Gillis, P.-A. Absil, and F.~Glineur, ``Two algorithms for
  orthogonal nonnegative matrix factorization with application to clustering,''
  \emph{Neurocomputing}, vol. 141, pp. 15--25, 2014.

\bibitem{bertsekas1999nonlinear}
D.~P. Bertsekas, \emph{Nonlinear programming}.\hskip 1em plus 0.5em minus
  0.4em\relax Athena scientific Belmont, 1999.

\bibitem{bengio2007greedy}
Y.~Bengio, P.~Lamblin, D.~Popovici, and H.~Larochelle, ``Greedy layer-wise
  training of deep networks,'' in \emph{Advances in neural information
  processing systems}, 2007, pp. 153--160.

\bibitem{ding2010convex}
C.~H. Ding, T.~Li, and M.~I. Jordan, ``Convex and semi-nonnegative matrix
  factorizations,'' \emph{IEEE transactions on pattern analysis and machine
  intelligence}, vol.~32, no.~1, pp. 45--55, 2010.

\bibitem{lee2001algorithms}
D.~D. Lee and H.~S. Seung, ``Algorithms for non-negative matrix
  factorization,'' in \emph{Advances in neural information processing systems},
  2001, pp. 556--562.

\bibitem{guan2012nenmf}
N.~Guan, D.~Tao, Z.~Luo, and B.~Yuan, ``Nenmf: An optimal gradient method for
  nonnegative matrix factorization,'' \emph{IEEE Transactions on Signal
  Processing}, vol.~60, no.~6, pp. 2882--2898, 2012.

\bibitem{cai2011graph}
D.~Cai, X.~He, J.~Han, and T.~S. Huang, ``Graph regularized nonnegative matrix
  factorization for data representation,'' \emph{IEEE Transactions on Pattern
  Analysis and Machine Intelligence}, vol.~33, no.~8, pp. 1548--1560, 2011.

\bibitem{sim2002cmu}
T.~Sim, S.~Baker, and M.~Bsat, ``The cmu pose, illumination, and expression
  (pie) database,'' in \emph{Automatic Face and Gesture Recognition, 2002.
  Proceedings. Fifth IEEE International Conference on}.\hskip 1em plus 0.5em
  minus 0.4em\relax IEEE, 2002, pp. 53--58.

\bibitem{KCLee05}
K.~Lee, J.~Ho, and D.~Kriegman, ``Acquiring linear subspaces for face
  recognition under variable lighting,'' \emph{IEEE Trans. Pattern Anal. Mach.
  Intelligence}, vol.~27, no.~5, pp. 684--698, 2005.

\end{thebibliography}

\end{document}